%% file: main.tex
\documentclass[a4paper]{cas-sc}

\usepackage[authoryear,longnamesfirst]{natbib}
\usepackage{graphicx,lipsum,subfigure}
\usepackage{tabularx} 
\usepackage{fancyhdr,graphicx,amsmath,amssymb}
\usepackage{bbold}
\usepackage[ruled,vlined, linesnumbered]{algorithm2e}
\usepackage{float}

\def\tsc#1{\csdef{#1}{\textsc{\lowercase{#1}}\xspace}}
\tsc{WGM}
\tsc{QE}
\begin{document}
\let\WriteBookmarks\relax
\def\floatpagepagefraction{1}
\def\textpagefraction{.001}

\shorttitle{ViSketch-GPT }    

\shortauthors{Giulio Federico, Giuseppe Amato, Fabio Carrara, Claudio Gennaro, Marco \text{Di Benedetto}}  

\title [mode = title]{ViSketch-GPT: Collaborative Multi-Scale Feature Extraction for Sketch Recognition and Generation} 

\tnotemark[1] 

\tnotetext[1]{} 


\author[1,2]{Giulio Federico}[orcid=0009-0005-0879-5631]

\cormark[1]

\ead{giulio.federico@isti.cnr.it}

\author[1]{Giuseppe Amato}[orcid=0000-0003-0171-4315]

% Email id of the second author
\ead{giuseppe.amato@isti.cnr.it}

\author[1]{Fabio Carrara}[orcid=0000-0001-5014-5089]

% Email id of the second author
\ead{fabio.carrara@isti.cnr.it}

\author[1]{Claudio Gennaro}[orcid=0000-0002-3715-149X]

% Email id of the second author
\ead{claudio.gennaro@isti.cnr.it}

\author[1]{Marco \text{Di Benedetto}}[orcid=0000-0001-5781-7060]

% Email id of the second author
\ead{marco.dibenedetto@isti.cnr.it}

\affiliation[1]{organization={ Institute of Information Science and Technologies (ISTI-CNR)},
            addressline={ Via Giuseppe Moruzzi 1}, 
            city={Pisa},
            postcode={56127}, 
            state={PI},
            country={Italy}}

\affiliation[2]{organization={University of Pisa},
            city={Pisa},
            postcode={56127}, 
            state={PI},
            country={Italy}}

\cortext[1]{Corresponding author}

\fntext[1]{}


\begin{abstract}
Understanding the nature of human sketches is challenging because of the wide variation in how they are created. Recognizing complex structural patterns improves both the accuracy in recognizing sketches and the fidelity of the generated sketches. In this work, we introduce \textbf{ViSketch-GPT}, a novel algorithm designed to address these challenges through a multi-scale context extraction approach. The model captures intricate details at multiple scales and combines them using an ensemble-like mechanism, where the extracted features work collaboratively to enhance the recognition and generation of key details crucial for classification and generation tasks.

The effectiveness of ViSketch-GPT is validated through extensive experiments on the QuickDraw dataset. Our model \textit{establishes a new benchmark}, significantly \textbf{outperforming existing methods} in both classification and generation tasks, with substantial improvements in accuracy and the fidelity of generated sketches.

The proposed algorithm offers a robust framework for understanding complex structures by extracting features that collaborate to recognize intricate details, enhancing the understanding of structures like sketches and making it a versatile tool for various applications in computer vision and machine learning.

\end{abstract}


%\nocite{*}

% Keywords

\maketitle

\input{doc/ack}
\input{doc/introduction}

\input{doc/related}
\input{doc/methodology}
\input{doc/experiments}

\input{doc/conclusion}
\input{doc/declaration_genai}

% Use the style of numbering in square brackets.
% If nothing is used, default style will be taken.
%\begin{enumerate}[a)]
%\item 
%\item 
%\item 
%\end{enumerate}  

%\begin{itemize}
%\item 
%\item 
%\item 
%\end{itemize}  

%\begin{description}
%\item[]
%\item[] 
%\item[] 
%\end{description}  

% Uncomment and use as the case may be
%\begin{theorem} 
%\end{theorem}

% Uncomment and use as the case may be
%\begin{lemma} 
%\end{lemma}


\bibliographystyle{cas-model2-names}

\bibliography{cas-refs}



\end{document}

%% file: doc/ack.tex
\section*{Acknowledgement}
This work has received financial support by the Horizon Europe Research \& Innovation Programme under Grant agreement N. 101092612 (Social and hUman ceNtered XR - SUN project) and by project "Italian Strengthening of ESFRI RI RESILIENCE" (ITSERR) funded by the European Union under the NextGenerationEU funding scheme (CUP:B53C22001770006).

%% file: doc/introduction.tex
\section{Introduction}\label{introduction}

Recognizing patterns of complex structures is fundamental for both the recognition and generation of visual content. However it becomes particularly challenging in the domain of human sketches where there is no single way to draw, but rather a variety of styles and representations for the same entities. Current approaches struggle to recognize and capture complex patterns, as they often fail to identify the intricate details that distinguish one entity from another, which are essential for accurate recognition and would enable better generation even when treating sketches as vector representations and leveraging NLP techniques to capture intricate temporal and structural dependencies. Vector sketches are inherently represented as ordered sequences of strokes, where each stroke is defined by a series of connected points. This sequential structure aligns naturally with NLP methodologies, which are designed to handle ordered, context-dependent data.

Models like SketchRNN (\cite{ha2017neuralrepresentationsketchdrawings}) have demonstrated the importance of leveraging sequential stroke information for sketch generation, employing recurrent architectures (RNNs) to predict the next stroke based on previous ones. Building on this, Sketch-BERT (\cite{lin2020sketchbertlearningsketchbidirectional}) extended the approach by adapting language modeling techniques such as BERT (\cite{devlin-etal-2019-bert}) to the sketch domain. This enabled not only sketch generation but also improved recognition and retrieval, capitalizing on the transformer's ability to capture bidirectional context. By treating sketches as sequences of visual "tokens," these approaches bridge the gap between textual and visual data, unlocking new possibilities for understanding and generating free-hand sketches.
\newline
In this work, we present a novel framework that redefines how context is extracted and utilized for both recognition and generation. By decomposing the sketch into smaller patches using the quadtree technique and employing a multi-level context extraction mechanism for each patch, ViSketch-GPT captures contextual information at different scales, allowing each patch to be more accurately characterized. The collaborative integration of these features enables the model to capture intricate details, which are essential for precise recognition and significantly enhance the generation process by maintaining structural coherence.

To evaluate the performance of ViSketch-GPT, we conducted experiments to assess its ability to recognize sketches and the fidelity of the generated sketches. Our results demonstrate that the model \textbf{significantly outperforms the state of the art} in classification, surpassing existing methods in both Top-1 and Top-3 accuracy, and demonstrating superior fidelity in sketch generation, as shown by the classifier's ability to accurately recognize the new generated sketches.

Our contributions are as follows: 
\begin{enumerate}
    \item We introduce a new methodology for capturing intricate details through the collaboration of multi-scale features, enhancing both sketch recognition and generation.
    
    \item We evaluate the methodology through extensive experiments on the QuickDraw dataset, demonstrating its \textit{superior performance compared to state-of-the-art methods} in both sketch recognition and generation.

    \item We have optimized our approach to handling \textit{sparse data} through a representation that mitigates potential issues during the generation phase.
\end{enumerate}

%% file: doc/related.tex
\section{Related works}\label{related}

Since the introduction of Sketch-a-Net in 2015 (\cite{yu2015sketchanetbeatshumans}) as a CNN-based model capable of generating free-hand sketches, significant advancements have been made in this domain in terms of architectures, representations, and datasets. In 2017, Google released QuickDraw, a large-scale sketch dataset comprising over 50 million sketches collected from players worldwide and SketchRNN (\cite{ha2017neuralrepresentationsketchdrawings}) was introduced as an RNN-based deep Variational Autoencoder (VAE) capable of generating diverse sketches. Subsequent developments have focused on retrieval methods (\cite{xu2018sketchmatedeephashingmillionscale}), recognition approaches (\cite{xu2021multigraphtransformerfreehandsketch, hu2018sketchaclassifiersketchbasedphotoclassifier}), and abstraction techniques aimed at simplifying sketches while preserving their recognizability (\cite{muhammad2018learningdeepsketchabstraction}), among others. Additionally, numerous new datasets have emerged, spanning both uni-modal and multi-modal domains, as well as varying levels of granularity (coarse- vs. fine-grained). Uni-modal datasets primarily support tasks such as recognition, retrieval, segmentation, and generation, whereas multi-modal datasets associate sketches with other modalities, including natural images, 3D models, and textual descriptions. Coarse-grained datasets (\cite{10.1145/2185520.2185540, ha2017neuralrepresentationsketchdrawings}) provide more general sketch representations, while fine-grained datasets offer higher levels of detail (\cite{7780462}).

Free-hand sketch tasks can be categorized into \textit{uni-modal} and \textit{multi-modal} tasks based on the type of data involved.  

Uni-modal tasks include \textit{recognition, retrieval, segmentation}, and \textit{generation}. Recognition aims to predict the class of a given sketch (\cite{7780494, 7153606, Zhang2016DeepNN,lin2020sketchbertlearningsketchbidirectional, 7327196, 10.5555/3015812.3015979, 10.1145/2964284.2973828, 8694004, 10.1145/3394171.3413810}). Retrieval focuses on using a query sketch to retrieve similar samples from a dataset or collection (\cite{lin2020sketchbertlearningsketchbidirectional,7351724, xu2018sketchmatedeephashingmillionscale, creswell2016adversarial}). This is a particularly challenging task since traditional feature extraction methods (e.g., SIFT (\cite{Lowe2004DistinctiveIF})) are ineffective due to the difficulty of identifying repeatable feature points across sketches drawn in diverse human styles. Segmentation involves the semantic partitioning of sketches and while conventional segmentation models for natural images could be adapted, the sequential nature of sketches has led to the development of dedicated models tailored to this task \cite{ Sketchsegnet, 8766108, SPFusionNet, https://doi.org/10.1111/cgf.13365, Kaiyrbekov_2020, 10.1145/3450284, Wang_2024_CVPR}).  

Deep learning-based approaches (\cite{ha2017neuralrepresentationsketchdrawings, 10.1609/aaai.v33i01.33012564, 8453841, li2020sketchman, ge2021creativesketchgeneration, das2020beziersketchgenerativemodelscalable, 10.1145/3414685.3417840, ribeiro2020sketchformertransformerbasedrepresentationsketched, das2021cloud2curvegenerationvectorizationparametric, tiwari2024sketchgptautoregressivemodelingsketch}) have significantly outperformed traditional sketch generation methods. Sketch generation has numerous practical applications, including synthesizing new sketches, assisting artists in streamlining their design process, and reconstructing corrupted or incomplete sketches.  

A pioneering model in this domain is \text{SketchRNN} (\cite{ha2017neuralrepresentationsketchdrawings}), which remains one of the state-of-the-art approaches for sketch-based abstraction and generalization. SketchRNN is a recurrent neural network-based generative model designed for both conditional and unconditional vector sketch generation. It was trained on QuickDraw, a large-scale dataset of vector drawings collected from "Quick, Draw!", an online game where players were asked to sketch objects from a given category within 20 seconds. The dataset includes hundreds of object categories, each containing 70 training samples, along with 2.5K validation and test samples.  
The network is formally a Sequence-to-Sequence Variational Autoencoder (VAE), where the encoder is a bidirectional RNN that takes as input the sketch (the sequence defining it) and outputs a latent vector (the concatenation of the two hidden states obtained from the bidirectional RNNs). This latent vector is then transformed via a fully connected layer into a vector representing the mean and standard deviation, which are used to sample a latent vector. This sampled latent vector is provided as input to a decoder (an autoregressive RNN), which samples the subsequent strokes of the sketch.

SketchBERT (\cite{lin2020sketchbertlearningsketchbidirectional}) is a model based on BERT (Bidirectional Encoder Representations from Transformers \cite{devlin-etal-2019-bert}) adapted for handling free-hand sketches. It leverages BERT’s ability to capture contextual relationships between elements in a sequence to analyze and interpret sketches, treating them as temporal sequences of pen strokes. SketchBERT is designed for sketch recognition and generation tasks, utilizing a pre-trained representation to enhance performance across various computer vision and sketch generation tasks while maintaining high efficiency in the context of unstructured sequential data like sketches.

AI-Sketcher (\cite{10.1609/aaai.v33i01.33012564}) proposes an enhancement of Sketch-RNN for handling multi-class generation, also based on a VAE generative model. They evaluated their method on a single dataset: FaceX. This dataset contains 5 million sketches of male and female facial expressions, and, unlike QuickDraw, the sketches were created by professionals. 

VASkeGAN (\cite{balasubramanian2019teaching}) combines a Variational Autoencoder (VAE) with a Generative Adversarial Network (GAN) to leverage the strengths of both models. The VAE is used to obtain an efficient representation of the data, while the GAN is employed to generate high-quality images. The goal is to produce visually appealing sketches, benefiting from both the compression capabilities of the VAE and the realistic generation abilities of the GAN. Additionally, a new metric called \textit{SkeScore} was introduced; however, it is only applicable to vector-based generations.

SketchGPT (\cite{tiwari2024sketchgptautoregressivemodelingsketch}) employs a sequence-to-sequence autoregressive model for sketch generation and completion by mapping complex sketches into simplified sequences of abstract primitives by leveraging the next token prediction objective strategy to understand sketch patterns, facilitating the creation and completion of drawings and also categorizing them accurately.

%% file: doc/methodology.tex
\section{Methodology}

\begin{figure}[h]
    \centering
    \includegraphics[width=0.4\textwidth]{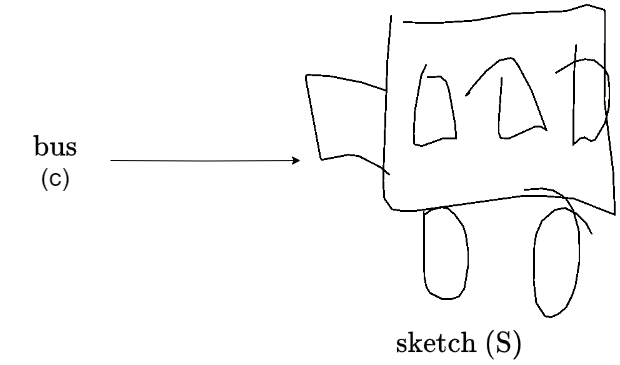}
    \caption{An example of the task we aim to tackle. Starting from the class label, we aim to generate a sketch belonging to that class.}
    \label{fig:task}
\end{figure}

We formulate the problem as follows. We want to train a Neural Network that, given a specific class, is able to generate an image containing a sketch which belongs to the class.  %\newline
More formally, our objective is to learn the conditional distribution \( p(S | c) \) so that, given the class label \( c \in \mathbb{Z}^+ \), it is possible to generate \( S \in \mathbb{R}^{H \times W} \), representing a sketch of \( c\).
 An example of such a task is shown in Figure \ref{fig:task}.

We split the generation process into \textbf{two stages}: i) pure generation and ii) refinement generation.  \newline

In the \textbf{pure generation stage}, we tackle the task of modeling the distribution \( p(S | c) \) in a much smaller resolution space than the desired one. Specifically, if \( H, W \) are the desired dimensions of the sketches, we choose \( H', W' \) such that \( H' \ll H, W' \ll W \). More formally, this first stage can be defined as the process of inferring the conditional distribution:

$$
p(S' | c) \quad \text{where } c\in \mathbb{Z}^+ \text{and } S'\in \mathbb{R}^{H' \times W'}
$$

This simplifies and significantly accelerates the training process on complex shapes like sketches, as working in a lower-dimensional space reduces the complexity of modeling the distribution. Instead of accounting for all highly variable human-specific details, the model can focus on capturing the essential structural characteristics representative of the class, making learning more efficient. To model this distribution, we opted for \textit{diffusion model theory} (\cite{ho2020denoising}). \newline

In the \textbf{generative refinement} stage, our goal is to \textit{restore the details} that were lost due to the low resolution of \(S'\). For a single \(S'\), there can exist multiple versions of \(S\) that contain different details, all statistically plausible.

The objective of the second stage is to learn how to generate a plausible higher resolution version of \(S'\), namely \(S\), by inferring the following conditional distribution:

$$
p(S | S',c) \quad \text{where } c\in \mathbb{Z}^+, S\in \mathbb{R}^{H \times W} \text{and } S'\in \mathbb{R}^{H' \times W'}
$$

\begin{figure}[h]
    \centering
    \includegraphics[width=0.48\textwidth]{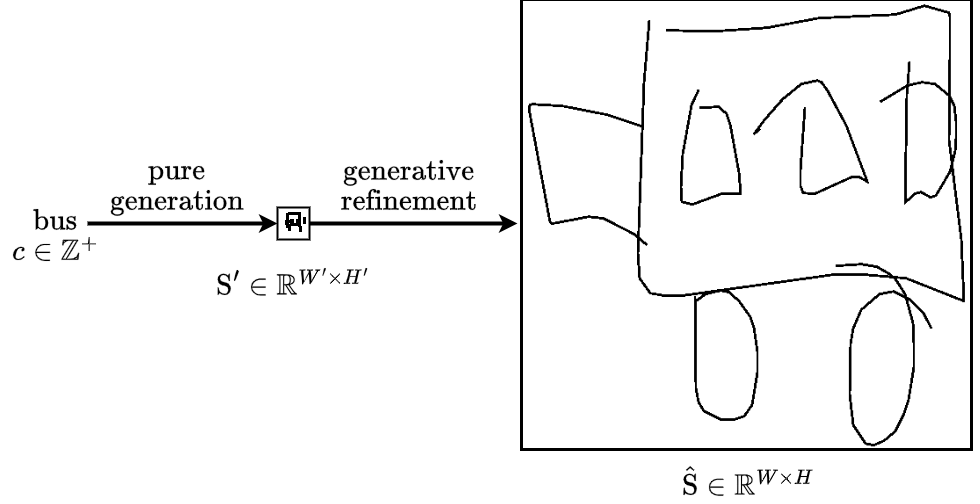}
    \caption{
Overview of the two stages: the first stage operates at a very low resolution to simplify and accelerate modeling; the second stage generates plausible details in a scalable manner.}
    \label{fig:overview}
\end{figure}

An overview of the two stages is shown in Figure \ref{fig:overview}. 

The way in which this distribution is modeled in the second stage is at the core of the algorithm. Further details are provided in the following sections. \newline

\subsection{Spatial Context Extraction for Scalable Refinement}

The goal of the \textbf{generative refinement} stage is to to model the distribution \( p(S | S',c) \). In this way, it is possible to plausibly generate a higher resolution version consistent with \(S'\) (the output of the previous stage) and with the class \(c\) by generatively restoring its details.

A naive approach would be to condition a generative network on \(S'\) and \(c\) to generate the entire \(S\) at its original resolution. Another approach involves dividing \(S\) into patches and performing super-resolution on the independent patches. These techniques may fail to fully capture the details we want to restore, let alone generate patches independently can lead to inconsistencies between adjacent patches previously generated, and generating a large number of patches—especially for sparse data—could be avoided to speed up the generation.

The primary goal is therefore to model the distribution ensuring the coherent reconstruction of missing details while avoiding minimizing redundant generation.

For this reason, we propose a \textbf{novel pipeline}.

Given a ground-truth sketch \(S\) of the class \(c\), and the low-resolution sketch \(S'\), generated in Stage 1, the training process for the generative refinement performs preliminary steps to allow working effectively at patch level.

\begin{figure}[h]
    \centering
    \includegraphics[width=0.48\textwidth]{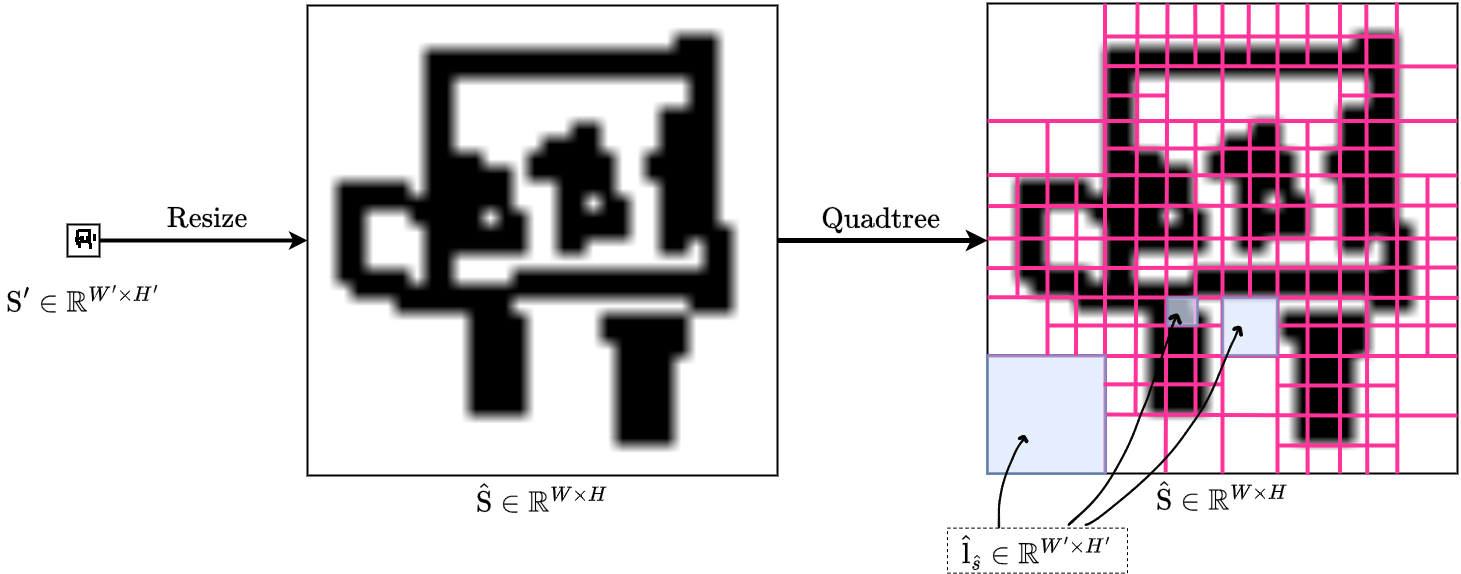}
    \caption{
First step of the generative refinement pipeline. Given the output of stage 1, \(S'\) is resized to the original resolution \(\hat{S}\) and the quadtree is computed.}
    \label{fig:resize_and_quadtree}
\end{figure}

The \textbf{first step}, performs a trivial \textit{resize} operation on \(S'\), to adapt it to the desired target resolution. We will refer to this resized version as \(\hat{S}\). Then, we build a quadtree by recursively partition \(\hat{S}\) only if a certain tile contains significant information or until each leaf reaches the same resolution used in the first stage: \(W'\times H'\) (Figure \ref{fig:resize_and_quadtree}). Note that when we build the quadtree,  we ensure that \textit{all leaves, regardless of their level, have the same resolution} (\(\hat{l}_{\hat{s}} \in \mathbb{R}^{W'\times H'}\)).

\begin{figure}[h]
    \centering
    \includegraphics[width=0.48\textwidth]{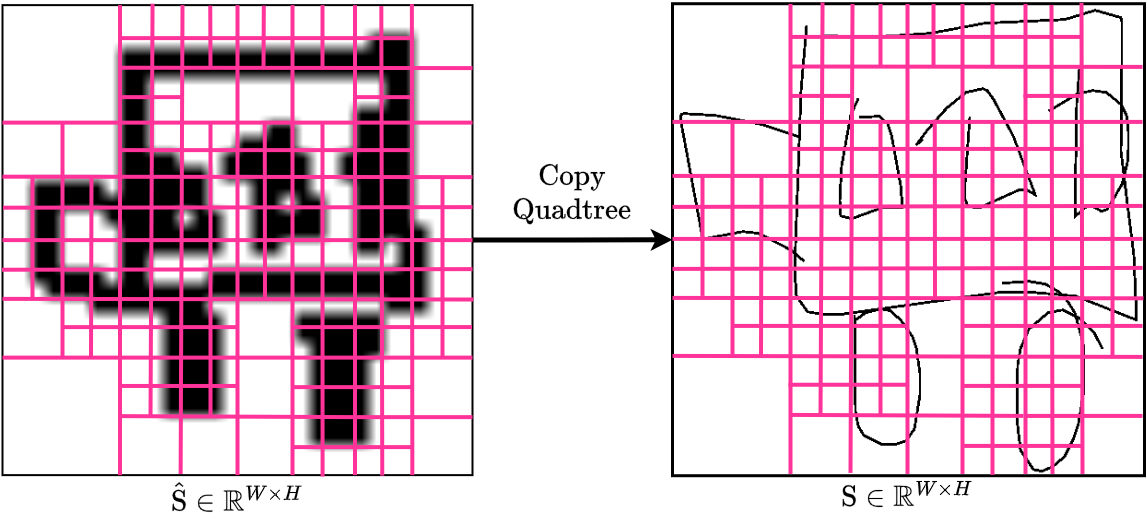}
    \caption{
The second step of the generative refinement pipeline. Copy the quadtree of \(\hat{S}\) into\(S\).}
    \label{fig:copy_quadtree}
\end{figure}

The \textbf{second step}, \textit{copies the quadtree}, obtained in previous step, onto the ground-truth sketch \(S\) (Figure \ref{fig:copy_quadtree}).

We will thus model the distribution \(p(S | S',c)\) as \(p(S | \hat{S},c)\), or equivalently, in terms of the leaves of the quadtree:

$$p(S | \hat{S},c)=p\left(\{l_{s}\} | \{\hat{l}_{\hat{s}}\},c\right)$$

which, by the property of joint probability, we can equivalently write as:

\begin{equation}
\label{eq:eq1}
\begin{aligned}
p\left(\{l_{s}\} | \{\hat{l}_{\hat{s}}\},c\right) &= p\left(l_{s}^{(1)},...,l_{s}^{(L)} | \hat{l}_{\hat{s}}^{(1)},...,\hat{l}_{\hat{s}}^{(L)},c\right) \\
&= p\left(l_{s}^{(1)} | \{\hat{l}_{\hat{s}}\},c\right) \cdot p\left(l_{s}^{(2)} | l_{s}^{(1)},\hat{l}_{\hat{s}}^{(2)},...,\hat{l}_{\hat{s}}^{(L)},c\right)...\\& ...\\
&\cdot p\left(l_{s}^{(i)} | l_{s}^{(1)},...,l_{s}^{(i-1)},\hat{l}_{\hat{s}}^{(i)},...,\hat{l}_{\hat{s}}^{(L)},c\right)\\&
..\\
&\cdot p\left(l_{s}^{(L)} | l_{s}^{(1)},...,l_{s}^{(L-1)},\hat{l}_{\hat{s}}^{(L)},c\right) \\
&=\prod_{i=1}^{L} p\left(l_{s}^{(i)} | l_{s}^{(1)},...,l_{s}^{(i-1)},\hat{l}_{\hat{s}}^{(i)},...,\hat{l}_{\hat{s}}^{(L)},c\right)
\end{aligned}
\end{equation}

where \(L\) is the number of leaves.

Each individual distribution is conditioned on the leaves of \(\hat{S}\) but also on all the previously generated leaves of \(S\), which will replace the corresponding ones in \(\hat{S}\). This will enable a \textit{coherent and aware generation} of what has been previously generated.

\begin{figure*}[t]
    \centering
    \includegraphics[width=\textwidth]{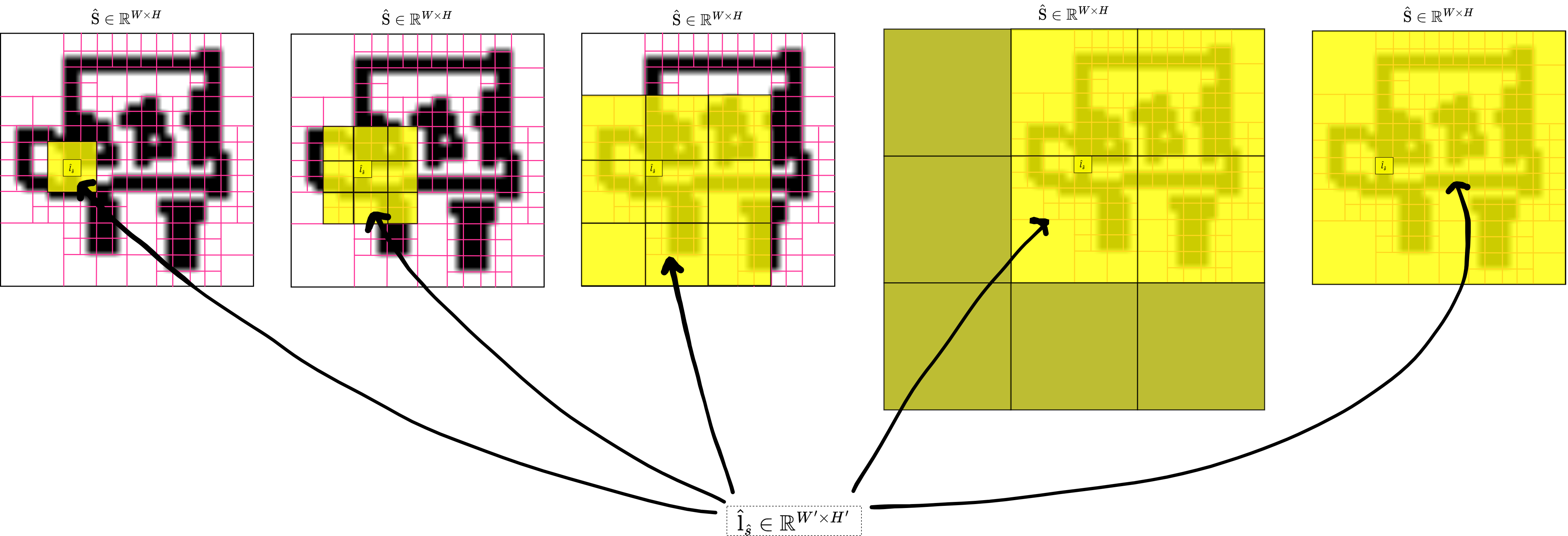}
    \caption{Process of creating the context of a leaf. Starting from the target leaf, the 3x3 tiles around it are taken with the leaf in the center. The same is done with the parent of the leaf until we reach the root itself. Each tile, regardless of the level, has the same resolution.}
    \label{fig:neighbors}
\end{figure*}

Note that the formulation (\ref{eq:eq1}) is quite naive and not scalable, as the generation of a single leaf \textit{needs all the initial and previously generated leaves}. For a better scalability, to generate a particular leaf we will use a much lighter condition that we will indicate as the \textit{context of the leaf to generate}. \newline
To this end, we define the \textbf{context of a leaf} \(\hat{l}_{\hat{s}^{(i)}}\) as the set of spatial neighbors of the leaf itself and all of its ancestors. Specifically, given the leaf \(\hat{l}_{\hat{s}^{(i)}}\), its spatial neighbors will correspond to the 3x3 grid of tiles where the leaf itself is at the center of the grid, and each tile has the same size as the leaf. The subsequent neighbors will be those of the 3x3 grid where the parent of the leaf is at the center, and each tile has the size of the parent. This process continues until the root itself is reached (the top-down view of \(\hat{S}\)). The entire process is illustrated in Figure \ref{fig:neighbors}. This context satisfies three fundamental properties:

\begin{itemize}
    \item The context of a leaf \(\hat{l}_{\hat{s}^{(i)}}\) must contain a lossless information about the surrounding pixels. This will help generate a leaf that is consistent with the values of the surrounding pixels.
    \item The context of a leaf \(\hat{l}_{\hat{s}^{(i)}}\) must contain progressively coarser information as it moves away from the target leaf.
    \item The context of a leaf must provide an overview of the current version of \(\hat{S}\).
\end{itemize}

Therefore, let \(\bar{S}_i\) denote the data at step \(i\) of the refinement, i.e.:

$$
    \bar{S}_i = \{l_{s}^{(1)},...,l_{s}^{(i-1)},\hat{l}_{\hat{s}}^{(i)},...,\hat{l}_{\hat{s}}^{(L)}\}
$$

and defining \(\aleph\left(\bar{l}_{\bar{s}^{(i)}}\right)\) as the context of a leaf in \(\bar{S}\), the formulation (\ref{eq:eq1}) becomes:

\begin{equation}
\label{eq:eq2}
p\left(\{l_{s}\} | \{\hat{l}_{\hat{s}}\},c\right)=\prod_{i=1}^{L} p\left(l_{s}^{(i)} | \aleph\left(\bar{l}_{\bar{s}^{(i)}}\right),c\right)
\end{equation}

\begin{figure}
    \centering
    \includegraphics[width=0.48\textwidth]{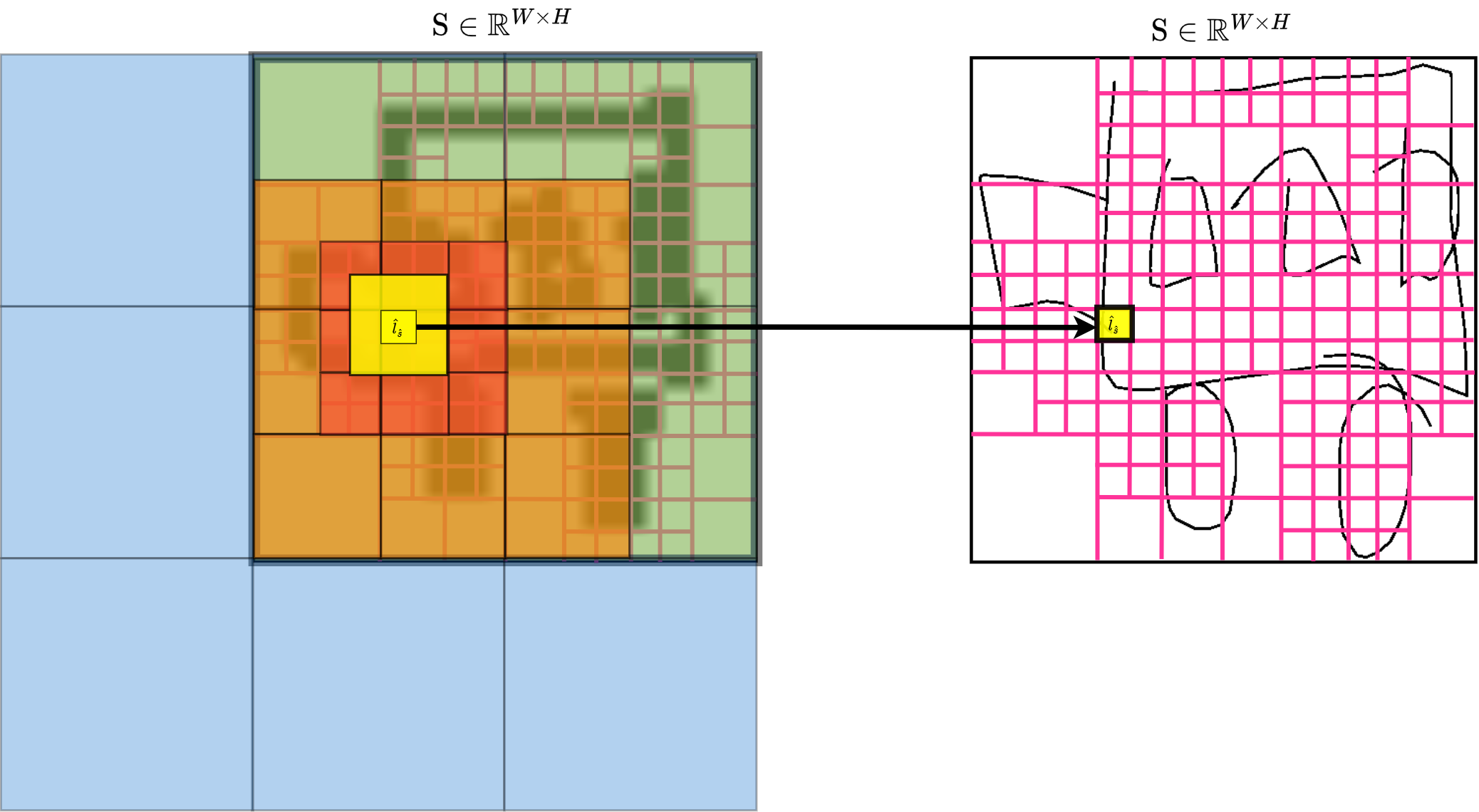}
    \caption{
To refine a leaf and make the approach more scalable, we consider its spatial context by including node's adjacent neighborhood (i.e., centered 3x3 grid) for each level up to the root. The nearby nodes in blue, which extend beyond the image, are referred to as \textit{dummy nodes} and have a fixed value.}
    \label{fig:train_refinement}
\end{figure}

During training, we will predict the generative refinement of the leaf \(l_{s}^{(i)}\) using the class \(c\) and the context \(\aleph(\cdot)\) of the leaf, as shown in Figure \ref{fig:train_refinement}.

\subsection{Generative refinement with autoregressive modeling}

To model each individual distribution in (\ref{eq:eq2}), we use a Transformer architecture (\cite{vaswani2023attentionneed}). Before training the Transformer \(\mathcal{T}_{\theta}\), we train a Vector Quantized Variational Autoencoder (VQ-VAE) (\cite{oord2018neuraldiscreterepresentationlearning}) to learn a discrete representation of the tiles at various resolutions they may have during the context creation process (Figure \ref{fig:neighbors}). This model acts as a \textbf{tokenizer}: each tile of size \(W' \times H'\) is encoded into a sequence of discrete indices belonging to a codebook. Once trained, the VQ-VAE is used to tokenize the refined patch into a sequence of discrete tokens, which are then used for training the Transformer.

The encoder is a vision encoder (\cite{dosovitskiy2021imageworth16x16words}) that extracts features from \(\bar{S}_i\) to enable the cross-attention mechanism with the decoder. However, unlike the classic ViT, we do not use all patches of \(\bar{S}_i\), as this would make the method unscalable. Instead, we select the context \(\aleph\left(\hat{l}_{\hat{s}^{(i)}}\right)\) of the leaf to be refined. A notable property of our methodology is that \textit{the sequence produced by context extraction is always of the same length}, regardless of the level at which the leaf is located. Indeed, if the octree has a maximum depth of \(D\), then the context \(\aleph\left(\hat{l}_{\hat{s}^{(i)}}\right)\) is a fixed sequence of length \(D \times 9 + 1\), where \(9\) represents the neighbors at each level, and the addition of \(1\) accounts for the contribution of the root. However, if a leaf belongs to an intermediate level \(0 < i < D\), its context still has the same length, but starting from a higher level, we use fixed and neutral values for the first elements of the sequence. \newline

\begin{figure*}[t]
    \centering
    \includegraphics[width=\textwidth]{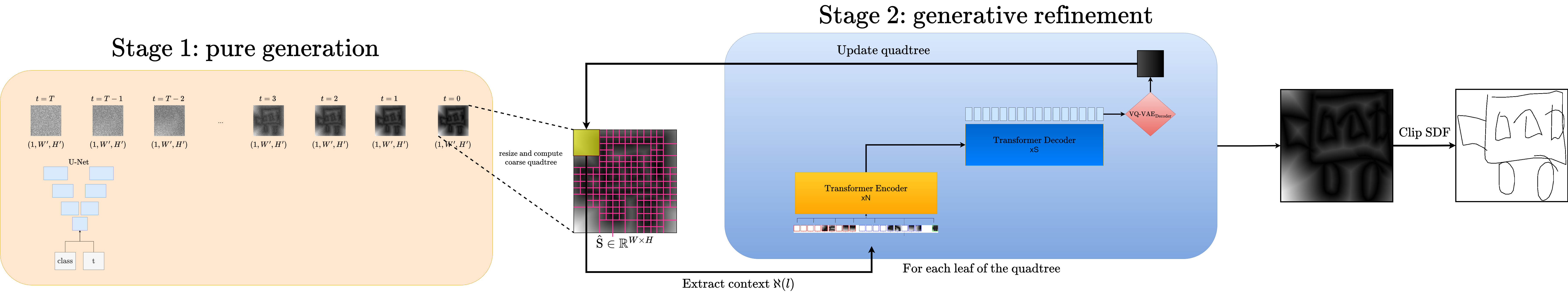}
    \caption{Architectural overview of the pure generation and generative refinement process. Given a class c, a low-resolution version of the SDF is generated, which is then scaled by a trivial resize to the desired resolution. The generative refinement stage, repeated for each leaf of the quadtree, restores the missing details. We end by clipping the SDF to get the data back in its sparse representation.}
    \label{fig:pipeline2}
\end{figure*}

The Transformer’s decoder outputs a sequence of probability distributions over the codebook indices, thus modeling the distribution of the refined leaf given the visual features extracted by the encoder. A visual representation of the architecture is shown in Figure \ref{fig:pipeline2}.

We train a VQ-VAE to compress each individual tile into a set of integer values \(\boldsymbol{z}\). To avoid low perplexity, and thus inefficient use of the codebook, we train the VQ-VAE not on the sparse data, but on a different representation that "intelligently fills" the empty spaces, preventing codebook collapse, which would introduce a strong bias in the transformer, leading to the prediction of repetitive sequences or overemphasizing certain tokens. This representation consists of calculating the \textbf{Signed Distance Fields (SDF)} of the sparse data. A SDF is a scalar field that represents the distance from a given point to the nearest surface. In our case, the surface corresponds to the stroke of the sketch. Positive values indicate points outside the stroke, while negative values indicate points inside it. This allows us to create a continuous representation of the shapes present in the sparse data, effectively filling in gaps and providing a more informative input for the VQ-VAE.

\begin{figure}[h]
    \centering
    \includegraphics[width=0.48\textwidth]{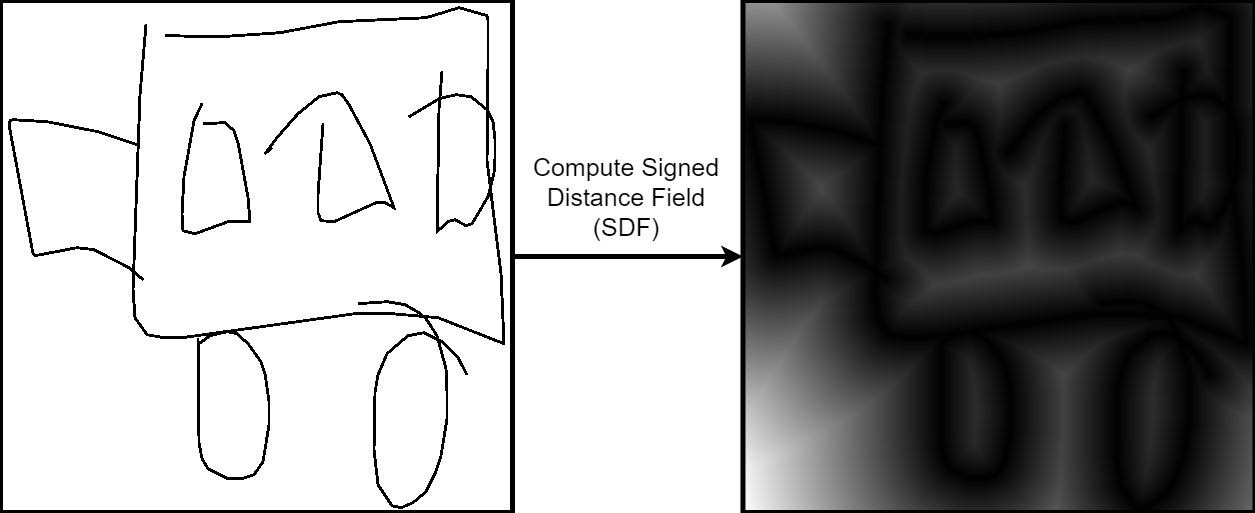}
    \caption{
Handling the Signed Distance Field (SDF) representation of sparse data helps VQ-VAE to have high perplexity (high codebook utilization) and thus avoid strong biases by the transformer in predicting certain classes.}
    \label{fig:trainrefinement}
\end{figure}

Specifically, the convolutional encoder performs a downsampling of each tile \(l\) to a smaller continuous spatial resolution:

$$
\mathcal{V}_\text{Encoder}(l)=\{v_1,v_2,....,v_K\}, \quad \text{where} \quad v_j\in R^{K,D}
$$

Subsequently, each continuous encoding \(v_i\) will be mapped to the nearest element of the codebook of vectors \(\mathbb{P} = \{p_i\}_{i=1}^Q \in \mathbb{R}^{Q,D}\), where \(Q\) is the total number of vectors in the codebook and \(D\) is the dimension of each vector. Therefore, in the end, each tile will be associated with a unique set of vectors:

$$
\mathbf{z}=\{q_1, q_2, ..., q_K\}, \quad \text{where} \quad q_i=\min_{p_j \in \mathbb{P}}|| v_i - p_j||
$$

whose codebook indices are chosen by the so-called quantizer \(\mathcal{V}_\text{Quantizer}(z)\).
Each tile is then decompressed using a convolutional decoder:

$$
\mathcal{V}_\text{Decoder}(\mathbf{z})=\hat{l}
$$

The entire model is trained end-to-end by minimizing the following loss (\cite{Corona-Figueroa_2023_ICCV}):

\begin{align}
\label{eq:6}
\mathcal{L}_{VQ} &= \mathcal{L}_{rec} \left(\hat{l}, l\right) + \mathcal{L}_{codebook}\left(v,z\right) \nonumber \\
&\quad + w_g \cdot \mathcal{L}_{generator}\left(\hat{l}\right) + w_d \cdot \mathcal{L}_{discriminator}\left(\hat{l}, l\right) \nonumber \\
&\quad + w_p \cdot \mathcal{L}_{perceptual}\left(\hat{l}, l\right)
\end{align}

The entire pipeline (generation + generative refinement) is shown in Figure \ref{fig:pipeline2}, while the pseudocode for training and inference is shown in algorithms (\ref{algo:stage2_training}) and (\ref{algo:stage2_inference}).

\begin{algorithm}
\small
\caption{Stage2: training}
\label{algo:stage2_training}
\SetAlgoNlRelativeSize{-1} 
\textbf{repeat}\\
\text{Choose random triplet \((S_0 \in \mathbb{R}^{W\times H}, S_0' \in \mathbb{R}^{W'\times H'}, c \in \mathbb{Z}^+)\)}\\
\text{ Resize \(S_0'\) to high resolution \(\rightarrow\) \(\hat{S_0}\in \mathbb{R}^{W\times H}\)} \\
\text{Compute the leaves of \(\hat{S_0}\) via quadtree \(\rightarrow\{\hat{l}_{\hat{s}_i}\}_{i=1}^L\)} \\
$l\sim Uniform({1...L})$\\
\text{Replace the leaves preceding \(l\) with the corresponding}
\text{tiles of \(S_0\)} $$\{  \hat{l}_{\hat{s}_i} | \hat{l}_{\hat{s}_i}=l_{s_i} \text{ for } i=1...(l-1) \}$$\\
\text{Compute target leaf context} \(\rightarrow \aleph\left(\hat{l}_{\hat{s}^{(i)}}\right)\) \\
\text{Tokenize the refined leaf using the learned codebook}
$$
       t= \mathcal{V}_\text{Quantizer}\left(\mathcal{V}_\text{Encoder}(l_{s_i})\right)
$$\\
\text{Compute the logits using Vision Encoder Decoder }
        $$
        \mathbf{g} = \mathcal{T}_{\theta}\left(\aleph\left(\hat{l}_{\hat{s}^{(i)}}\right),t, c\right)
        $$
\text{where \(\mathbf{g}\) is a sequence of K logits of lenght Q.} \text{Each logit is a vector as long as the VQ-VAE codebook,} \text{containing information on which codebook index is}
\text{most plausible to sample.} 
 \\

\text{Apply softmax for each one of the K logits}:
        $$
        \hat{y}_{k,j} = \frac{e^{g_{k,j}}}{\sum_{q'=1}^{Q} e^{g_{k,q'}}}, \quad \forall j = 1, \dots, Q
        $$

\text{Compute the loss function (cross-entropy)}
 $$
\mathcal{L} = - \frac{1}{K} \sum_{k=1}^{K} \sum_{q=1}^{Q} y_{k,q} \log \hat{y}_{k,q}
$$      

\text{Take gradient descent step on} 
        $$
        \nabla_{\theta} \mathcal{L}
        $$

\textbf{until }\text{convergence}

\end{algorithm}

\begin{algorithm}
\small
\caption{Stage2: inference}
\label{algo:stage2_inference}
\SetAlgoNlRelativeSize{-1}  
\textbf{Input:} \( S_0' \in \mathbb{R}^{W'\times H'} \), \( c \in \mathbb{Z}^+ \) \\
\text{ Resize \(S_0'\) to high resolution \(\rightarrow\) \(\hat{S_0}\in \mathbb{R}^{W\times H}\)}

\text{Compute the leaves of \( S_0 \) via quadtree} \(\rightarrow\{\hat{l}_{\hat{s}_i}\}_{i=1}^L\) \\

\textbf{for} \(\hat{l}_{\hat{s}_i}\) \textbf{in} \(\{\hat{l}_{\hat{s}_i}\}_{i=1}^L \) \textbf{do}: \\
\quad \text{Initialize token sequence} \( t = [] \)\\
\quad \textbf{for} \( d = 1 \) to \( K \) \textbf{do}: \\
\quad \quad \text{Compute the context of the leaf } \( \aleph(\hat{l}_{\hat{s}_i}) \) \\
\quad \quad \text{Compute logits using Vision Encoder Decoder}:
        $$
        \mathbf{g} = \mathcal{T}_{\theta}\left(\aleph(\hat{l}_{\hat{s}_i}), t, c\right)
        $$ \\
\quad \quad \text{Apply softmax to obtain probabilities:}
        $$
        \hat{y}_j = \frac{e^{g_j}}{\sum_{q'=1}^{Q} e^{g_q'}}, \quad \forall j = 1, \dots, Q
        $$ \\
\quad \quad \text{Sample token from predicted distribution:}  
        $$
        t_d \sim \text{Categorical}(\hat{y})
        $$ \\
\quad \quad \text{Append sampled token \( t_d \) to sequence \( t \)}

\quad \textbf{end for} \\
\quad \text{Decode token sequence into leaf patch:}  
$$
l_{s_d} = \mathcal{V}_\text{Decoder}(t)
$$ \\

\quad \text{Update quadtree}

\textbf{end for} \\

\end{algorithm}

%% file: doc/experiments.tex
\section{Experimental Validation}
To validate our methodology, we tested \textbf{ViSketch-GPT} on two types of tasks: sketch generation and classification.

\subsection{Dataset}
The proposed methodology is evaluated on the QuickDraw dataset (\cite{ha2017neuralrepresentationsketchdrawings}), which is a collection of sketches created for the Google application \textit{Quick, Draw!}, an online game in which users were asked to quickly draw, in less than 20 seconds, sketches related to specific categories. The dataset consists of 50 million sketches across 345 categories. 
Each sketch in QuickDraw is represented as a sequence of \textit{pen stroke actions}, defined by five elements:
\[
(\Delta_x, \Delta_y, p_1, p_2, p_3)
\]
where:
\begin{itemize}
    \item $\Delta_x, \Delta_y$ represent the offset from the previous point.
    \item $p_1$ is a binary value indicating whether the pen is touching the paper (1) or lifted (0), thereby determining whether a line is drawn to the current point.
    \item $p_2$ is a binary indicator specifying whether the pen is lifted after reaching the current point (1) or not (0).
    \item $p_3$ denotes whether the drawing sequence ends at the current point (1) or continues (0).
\end{itemize}

In order to use our methodology, we transform each sketch into an image, which we then convert it into an SDF.

We followed the pre-processing method and training split suggested by \cite{ha2017neuralrepresentationsketchdrawings}, where each class has 70K training samples, 2.5K validation, and 2.5K test samples in the QuickDraw dataset. In order to align with various competitors, we also simplified the sketches by applying the \textit{Ramer-Douglas-Peucker} (RDP) algorithm to handle sequences with a maximum length of 321.

\subsection{Results on Sketch Generation Task}

\sloppy For the sketch generation task, we followed the methodology proposed by SketchGPT (\cite{tiwari2024sketchgptautoregressivemodelingsketch}) to ensure a fair comparison with SketchRNN. Their approach quantifies the generative capability of a model by evaluating how well a classifier can distinguish generated sketches. 

To maintain a fair comparison, we selected the same seven classes from the QuickDraw dataset (bus, cat, elephant, flamingo, owl, pig and sheep) and trained a ResNet34 model to classify sketches from these categories. The model was trained until it reached a validation accuracy of 87.92\%. Subsequently, we generated 1,000 sketches per class, resulting in a total of 7,000 sketches, which were then fed into the previously trained ResNet34 to measure top-1 and top-3 accuracy.

\begin{figure*}[t]
    \centering
    \includegraphics[width=0.6\textwidth]{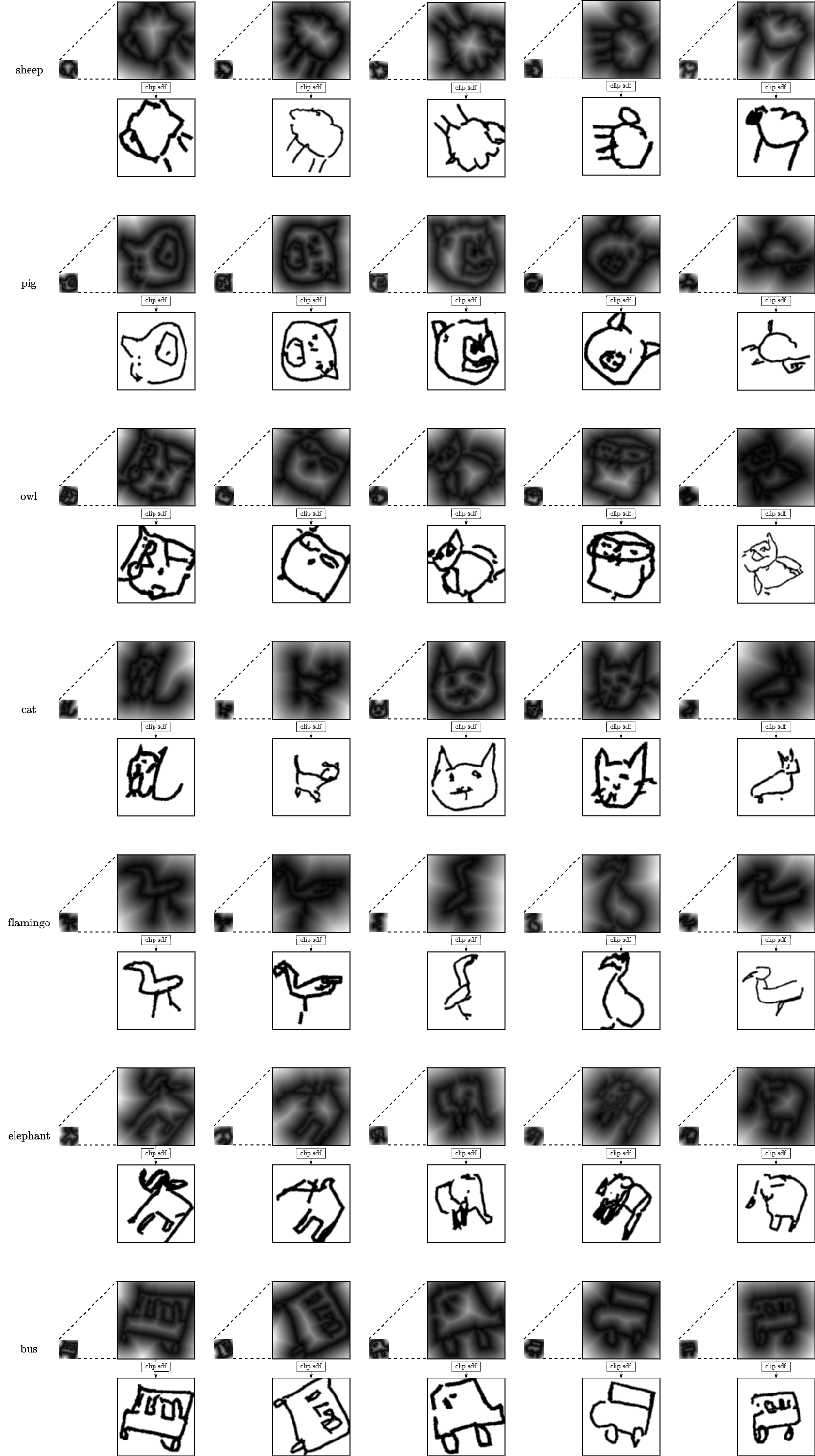}
    \caption{Some results of the generation task. Given a class, a low-resolution version of the SDF of the sketch is generated, which is then refined in a generative way and then clipped to obtain the final sketch.}
    \label{fig:gen_results}
\end{figure*}

Table \ref{tab:gen_sketch_comparison} presents the performance results of the models. Additionally, we report the results of our model both with and without the refinement stage. Our findings highlight the critical role of the use of SDF in the first stage and the use of this refinement stage in significantly improving generation quality, leading to a \textbf{substantial increase in classification accuracy compared to the state-of-the-art}. This improvement is likely due to the refinement process adding missing or unclear details, enabling the CNN to classify the generated sketches with higher confidence. Figure \ref{fig:gen_results} presents some examples of generated sketches for the target classes.

\begin{table}[ht]
    \centering
    \resizebox{0.7\columnwidth}{!}{ % Ridimensiona la tabella alla larghezza di una colonna
        \begin{tabular}{lcc} % l = left, c = center
            \toprule
            \textbf{Method} & \textbf{Top-1 Acc.} & \textbf{Top-3 Acc.} \\
            \midrule
            SketchRNN  & 44.6\% & 79.1\% \\
            SketchGPT    & 50.4\% & 81.7\% \\
            \textbf{ViSketch-GPT (our) (w/o Ref.)}  & \textbf{51.6}\% & \textbf{83.2}\% \\
            \textbf{ViSketch-GPT (our) (w Ref.)}  & \textbf{63.4}\% & \textbf{88.8}\% \\
            \bottomrule
        \end{tabular}
    }
    \caption{Performance comparison of different sketch generation models. The designations "w/o" and "w" indicate results without and with the refinement stage, respectively.}
    \label{tab:gen_sketch_comparison}
\end{table}

\subsection{Results on Sketch Classification Task}

\begin{figure*}[t]
    \centering
    \includegraphics[width=\textwidth]{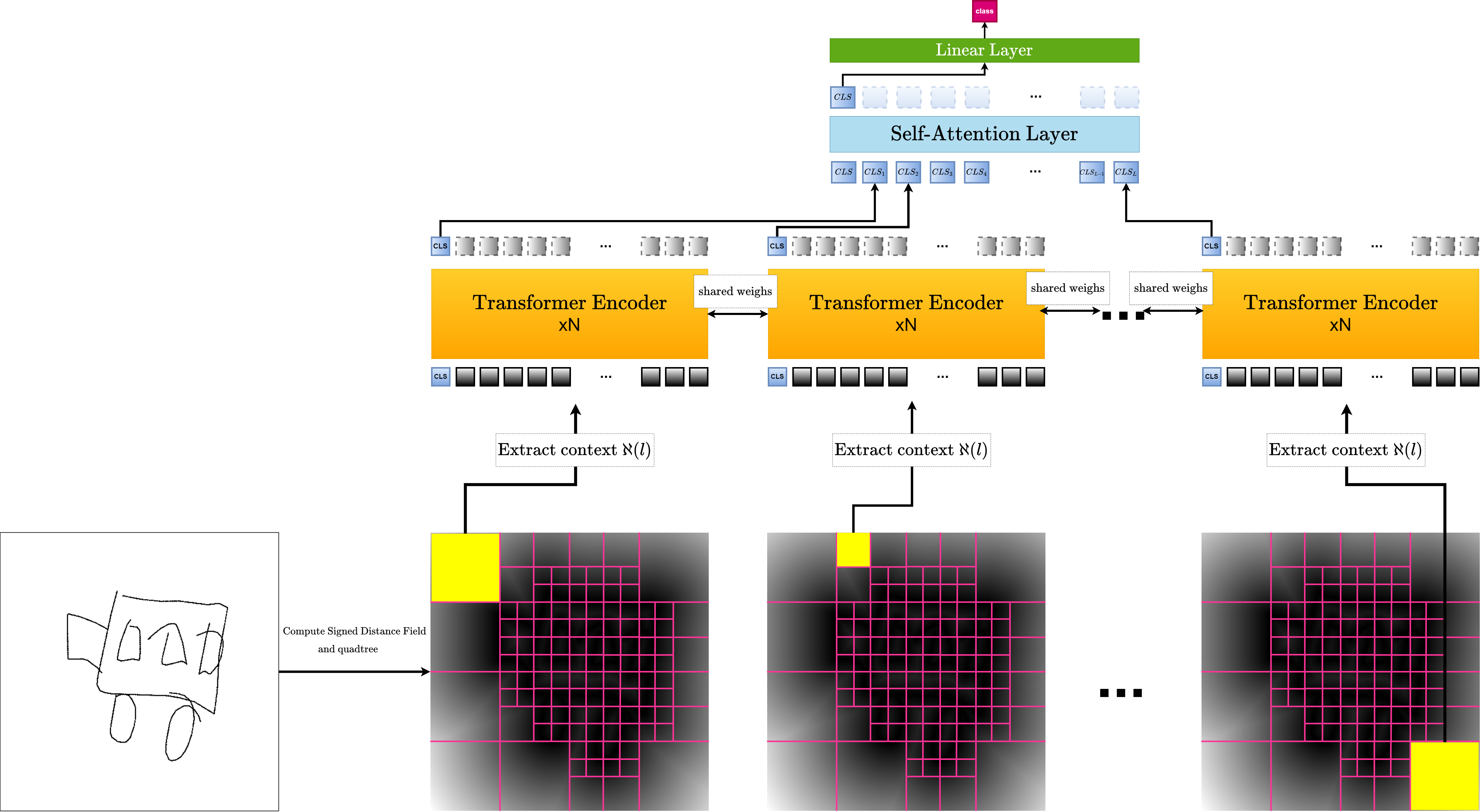}
    \caption{Architectural overview for the classification task. Once the encoder is pre-trained, it is used to extract context features for each leaf. Then the CLS is used as the vote of each leaf which is processed with the others for class prediction.}
    \label{fig:classification_architecture}
\end{figure*}

The Sketch Classification task is a typical task in which, given a sketch, one aims to classify it correctly into its corresponding category. We used 100 categories with 5K training samples, 2.5K validation samples, and 2.5K test samples from the QuickDraw dataset. We considered, as done previously by several competitors (\cite{tiwari2024sketchgptautoregressivemodelingsketch, lin2020sketchbertlearningsketchbidirectional}), evaluating whether the methodology could also be used as a feature extractor to train a classifier on top. In our case, given a sketch, we want to assess whether the features extracted via the proposed algorithm are significant enough to train a classifier that is more accurate than one trained using features extracted by other methods.

\textbf{Competitors} \newline
We conduct a comparative analysis of multiple baseline models:

\begin{enumerate}
    \item \textbf{HOG-SVM} (\cite{hog_svm}): A conventional approach that employs Histogram of Oriented Gradients (HOG) features combined with a Support Vector Machine (SVM) classifier to perform classification tasks.
    
    \item \textbf{Ensemble} (\cite{ensemble}): This model integrates various types of sketch features and is evaluated on classification tasks.
    
    \item \textbf{BiLSTM} (\cite{bilstm}): A three-layer bidirectional Long Short-Term Memory (BiLSTM) model is utilized to assess recognition and retrieval tasks on sequential sketch data. The hidden state dimensionality is set to 512.
    
    \item \textbf{Sketch-a-Net} (\cite{yu2015sketchanetbeatshumans}): A convolutional neural network specifically designed for sketch data.
    
    \item \textbf{DSSA} (\cite{dssa}): This approach extends Sketch-a-Net by incorporating an attention module and a high-order energy triplet loss function.
    
    \item \textbf{ResNet} (\cite{resnet}): A widely used residual neural network architecture in computer vision, primarily designed for image recognition tasks.
    
    \item \textbf{TC-Net} (\cite{tcnet}): A network based on DenseNet \cite{densenet}, applied to sketch-based image retrieval tasks. We utilize a pre-trained version for classification and retrieval experiments.
    
    \item \textbf{SketchRNN} (\cite{ha2017neuralrepresentationsketchdrawings}): A variational autoencoder that employs an LSTM-based encoder-decoder architecture for sketch generation. In our experiments, we adapt this model to evaluate the sketch gestalt task.

    \item \textbf{SketchBert} (\cite{lin2020sketchbertlearningsketchbidirectional}): A BERT-based model, adapted for free-hand sketches, uses BERT’s ability to capture contextual relationships in sequences to interpret sketches as temporal pen strokes.

    \item \textbf{SketchGPT} (\cite{tiwari2024sketchgptautoregressivemodelingsketch}): A model that employs a sequence-to-sequence autoregressive approach for sketch generation and completion, mapping complex sketches into simplified sequences of abstract primitives. By leveraging the next-token prediction strategy, it learns sketch patterns, enabling accurate drawing creation, completion, and categorization.

\end{enumerate}

The training and validation subsets are used for model training, while evaluations are conducted on the test set. \newline

\textbf{Implementation Details} \newline
The methodology was originally designed to refine a leaf \(\hat{l}_{\hat{s}_i}\) given its context \(\aleph(\hat{l}_{\hat{s}_i})\). However, to classify the entire sketch, we would need to process, after calculating the quadtree, \(L\) leaves of the sketch to obtain its class prediction. Therefore, we can interpret the classification task as an \textit{ensemble voting}, where each leaf, via its context \(\aleph(\cdot)\), provides its "proposal" for the class vote. We will therefore add a CLS token, which we will use after processing it with the encoder together with the leaf's context. Finally, we will use the \(L\) extracted CLS tokens, process them with a simple self-attention layer, and extract the final CLS token to make the prediction with a simple linear layer. Since pre-training tasks on unlabeled data in NLP have shown great potential in improving the performance of models on downstream tasks, following a similar approach to SketchBert with its "Sketch Gestalt" task (\cite{lin2020sketchbertlearningsketchbidirectional}), we pre-trained the VisionEncoderDecoder on the \textit{completion} task. Specifically, given a sketch, we calculate its quadtree, choose a random leaf, and mask it. Then, we train the model to predict the VQ-VAE codebook IDs, which, when decoded, represent the missing 2D tile. This task helps meaningfully initialize the encoder weights, which we then use to extract the leaf features for the classification task discussed earlier. \newline
Our architecture for pre-training has a depth of 6 in both the encoder and decoder, a hidden size of 256, and a number of heads equal to 8. For the pre-training phase, we use the same classes and sample size as SketchBert, i.e., 100 classes and 5K training examples, 2.5K for validation and testing, with an image resolution of 128 (leaf size of 32). An architectural overview is shown in Figure \ref{fig:classification_architecture}. \newline

\begin{table}[h]
    \centering
    \begin{tabularx}{\columnwidth}{Xcc}
        \toprule
        \textbf{Methods} & \textbf{Top-1 Acc.} & \textbf{Top-5 Acc.}\\
        \midrule
        HOG-SVM  & 56.13 & 78.34 \\
        Ensemble  & 66.98 & 89.32 \\
        Bi-LSTM  & 86.14 & 97.03 \\
        Sketch-a-Net  & 75.33 & 90.21 \\
        DSSA  & 79.47 & 92.41 \\
        ResNet18  & 83.97 & 95.98 \\
        ResNet50  & 86.03 & 97.06 \\
        TCNet  & 86.79 & 97.08 \\
        Sketch-BERT  (100 x 5K) & 85.82 & 97.31 \\
        Sketch-BERT  (200 x 5K) & 84.89 & 97.14 \\
        Sketch-BERT  (345 x 5K) & 85.73 & 97.31 \\
        Sketch-BERT  (345 x 70K) & 88.30 & 97.82 \\
        Sketch-GPT (50 x 5K) & 83.58 & 93.65 \\
        \textbf{ViSketch-GPT (our)} & \textbf{94.45} & \textbf{99.72} \\
        \bottomrule
    \end{tabularx}
    \caption{Comparison with state-of-the-art methods on sketch classification on the QuickDraw dataset. The notation (C x N) indicates the amount of pretraining done (i.e., with C classes and N samples for each class) before fine-tuning on the classes to be evaluated.}
    \label{tab:classification_quickdraw_results}
\end{table}

\textbf{Discussion}\newline
Table \ref{tab:classification_quickdraw_results} shows the top-1 and top-3 accuracy results on the QuickDraw dataset for both the competitors and our model. The results demonstrate that \textbf{our model significantly outperforms the state-of-the-art (SOTA)}, setting new performance benchmarks. Moreover, it is important to note that our pre-training was done on only 100 classes and 5K samples. Despite this minimal pre-training, our model outperformed others, including SketchBert, which was trained on up to 345 classes and 70K samples. This performance gain is likely attributed to the voting mechanism, which allows each leaf to contribute to the classification, helps capture those subtle details that ultimately make the difference in classifying the sketch rather than relying on a holistic approach.

%% file: doc/conclusion.tex
\section{Conclusion}\label{conclusion}

In this work, we tackled the challenge of generating and classifying human sketches, a complex task due to the high variability in strokes and representations. To overcome this difficulty, we introduced a novel algorithm capable of extracting significant multi-scale features, leveraging their collaboration to improve both sketch classification and generation. Our experiments demonstrate that this strategy achieves an unprecedented level of classification accuracy and generation quality, highlighting the effectiveness of our approach in capturing intricate details. These details enhance the classifier's confidence in understanding sketches and make the generator more aware of the characteristics that distinguish different sketches.  

While the results are encouraging, the proposed method has some limitations. In particular, our framework can be extended to higher-resolution sketches, and while this does not introduce memory issues, it remains an autoregressive approach that requires considerable time. A promising future direction is the exploration of parallelization techniques for leaf generation to accelerate the process and improve scalability.  

Ultimately, our work represents a step forward in understanding and modeling human sketches, opening new perspectives for applications in areas such as visual recognition and AI-assisted creativity.

%% file: doc/declaration_genai.tex
\section*{Declaration of generative AI and AI-assisted technologies in the writing process}

During the preparation of this work, the author(s) used ChatGPT, an AI language model developed by OpenAI, in order to correct grammatical errors. After using this tool, the author(s) reviewed and edited the content as needed and take(s) full responsibility for the content of the publication.

%% file: main.bbl
\begin{thebibliography}{49}
\expandafter\ifx\csname natexlab\endcsname\relax\def\natexlab#1{#1}\fi
\providecommand{\url}[1]{\texttt{#1}}
\providecommand{\href}[2]{#2}
\providecommand{\path}[1]{#1}
\providecommand{\DOIprefix}{doi:}
\providecommand{\ArXivprefix}{arXiv:}
\providecommand{\URLprefix}{URL: }
\providecommand{\Pubmedprefix}{pmid:}
\providecommand{\doi}[1]{\href{http://dx.doi.org/#1}{\path{#1}}}
\providecommand{\Pubmed}[1]{\href{pmid:#1}{\path{#1}}}
\providecommand{\bibinfo}[2]{#2}
\ifx\xfnm\relax \def\xfnm[#1]{\unskip,\space#1}\fi
%Type = Article
\bibitem[{Balasubramanian et~al.(2019)Balasubramanian, Balasubramanian et~al.}]{balasubramanian2019teaching}
\bibinfo{author}{Balasubramanian, S.}, \bibinfo{author}{Balasubramanian, V.N.}, et~al., \bibinfo{year}{2019}.
\newblock \bibinfo{title}{Teaching gans to sketch in vector format}.
\newblock \bibinfo{journal}{arXiv preprint arXiv:1904.03620} .
%Type = Inproceedings
\bibitem[{Ballester and Araujo(2016)}]{10.5555/3015812.3015979}
\bibinfo{author}{Ballester, P.}, \bibinfo{author}{Araujo, R.}, \bibinfo{year}{2016}.
\newblock \bibinfo{title}{On the performance of googlenet and alexnet applied to sketches}, in: \bibinfo{booktitle}{Proceedings of the AAAI conference on artificial intelligence}.
%Type = Article
\bibitem[{Bhunia et~al.(2020)Bhunia, Das, Muhammad, Yang, Hospedales, Xiang, Gryaditskaya and Song}]{10.1145/3414685.3417840}
\bibinfo{author}{Bhunia, A.K.}, \bibinfo{author}{Das, A.}, \bibinfo{author}{Muhammad, U.R.}, \bibinfo{author}{Yang, Y.}, \bibinfo{author}{Hospedales, T.M.}, \bibinfo{author}{Xiang, T.}, \bibinfo{author}{Gryaditskaya, Y.}, \bibinfo{author}{Song, Y.Z.}, \bibinfo{year}{2020}.
\newblock \bibinfo{title}{Pixelor: A competitive sketching ai agent. so you think you can sketch?}
\newblock \bibinfo{journal}{ACM Transactions on Graphics (TOG)} \bibinfo{volume}{39}, \bibinfo{pages}{1--15}.
%Type = Inproceedings
\bibitem[{Cao et~al.(2019)Cao, Yan, Shi and Chen}]{10.1609/aaai.v33i01.33012564}
\bibinfo{author}{Cao, N.}, \bibinfo{author}{Yan, X.}, \bibinfo{author}{Shi, Y.}, \bibinfo{author}{Chen, C.}, \bibinfo{year}{2019}.
\newblock \bibinfo{title}{Ai-sketcher: a deep generative model for producing high-quality sketches}, in: \bibinfo{booktitle}{Proceedings of the AAAI conference on artificial intelligence}, pp. \bibinfo{pages}{2564--2571}.
%Type = Inproceedings
\bibitem[{Corona-Figueroa et~al.(2023)Corona-Figueroa, Bond-Taylor, Bhowmik, Gaus, Breckon, Shum and Willcocks}]{Corona-Figueroa_2023_ICCV}
\bibinfo{author}{Corona-Figueroa, A.}, \bibinfo{author}{Bond-Taylor, S.}, \bibinfo{author}{Bhowmik, N.}, \bibinfo{author}{Gaus, Y.F.A.}, \bibinfo{author}{Breckon, T.P.}, \bibinfo{author}{Shum, H.P.H.}, \bibinfo{author}{Willcocks, C.G.}, \bibinfo{year}{2023}.
\newblock \bibinfo{title}{Unaligned 2d to 3d translation with conditional vector-quantized code diffusion using transformers}, in: \bibinfo{booktitle}{Proceedings of the IEEE/CVF International Conference on Computer Vision (ICCV)}, pp. \bibinfo{pages}{14585--14594}.
%Type = Inproceedings
\bibitem[{Creswell and Bharath(2016)}]{creswell2016adversarial}
\bibinfo{author}{Creswell, A.}, \bibinfo{author}{Bharath, A.A.}, \bibinfo{year}{2016}.
\newblock \bibinfo{title}{Adversarial training for sketch retrieval}, in: \bibinfo{booktitle}{Computer Vision--ECCV 2016 Workshops: Amsterdam, The Netherlands, October 8-10 and 15-16, 2016, Proceedings, Part I 14}, \bibinfo{organization}{Springer}. pp. \bibinfo{pages}{798--809}.
%Type = Inproceedings
\bibitem[{Das et~al.(2020)Das, Yang, Hospedales, Xiang and Song}]{das2020beziersketchgenerativemodelscalable}
\bibinfo{author}{Das, A.}, \bibinfo{author}{Yang, Y.}, \bibinfo{author}{Hospedales, T.}, \bibinfo{author}{Xiang, T.}, \bibinfo{author}{Song, Y.Z.}, \bibinfo{year}{2020}.
\newblock \bibinfo{title}{B{\'e}ziersketch: A generative model for scalable vector sketches}, in: \bibinfo{booktitle}{Computer Vision--ECCV 2020: 16th European Conference, Glasgow, UK, August 23--28, 2020, Proceedings, Part XXVI 16}, \bibinfo{organization}{Springer}. pp. \bibinfo{pages}{632--647}.
%Type = Inproceedings
\bibitem[{Das et~al.(2021)Das, Yang, Hospedales, Xiang and Song}]{das2021cloud2curvegenerationvectorizationparametric}
\bibinfo{author}{Das, A.}, \bibinfo{author}{Yang, Y.}, \bibinfo{author}{Hospedales, T.M.}, \bibinfo{author}{Xiang, T.}, \bibinfo{author}{Song, Y.Z.}, \bibinfo{year}{2021}.
\newblock \bibinfo{title}{Cloud2curve: Generation and vectorization of parametric sketches}, in: \bibinfo{booktitle}{Proceedings of the IEEE/CVF Conference on Computer Vision and Pattern Recognition}, pp. \bibinfo{pages}{7088--7097}.
%Type = Inproceedings
\bibitem[{Devlin et~al.(2019)Devlin, Chang, Lee and Toutanova}]{devlin-etal-2019-bert}
\bibinfo{author}{Devlin, J.}, \bibinfo{author}{Chang, M.W.}, \bibinfo{author}{Lee, K.}, \bibinfo{author}{Toutanova, K.}, \bibinfo{year}{2019}.
\newblock \bibinfo{title}{{BERT}: Pre-training of deep bidirectional transformers for language understanding}, in: \bibinfo{editor}{Burstein, J.}, \bibinfo{editor}{Doran, C.}, \bibinfo{editor}{Solorio, T.} (Eds.), \bibinfo{booktitle}{Proceedings of the 2019 Conference of the North {A}merican Chapter of the Association for Computational Linguistics: Human Language Technologies, Volume 1 (Long and Short Papers)}, \bibinfo{publisher}{Association for Computational Linguistics}, \bibinfo{address}{Minneapolis, Minnesota}. pp. \bibinfo{pages}{4171--4186}.
\newblock \URLprefix \url{https://aclanthology.org/N19-1423/}, \DOIprefix\doi{10.18653/v1/N19-1423}.
%Type = Article
\bibitem[{Dosovitskiy et~al.(2020)Dosovitskiy, Beyer, Kolesnikov, Weissenborn, Zhai, Unterthiner, Dehghani, Minderer, Heigold, Gelly et~al.}]{dosovitskiy2021imageworth16x16words}
\bibinfo{author}{Dosovitskiy, A.}, \bibinfo{author}{Beyer, L.}, \bibinfo{author}{Kolesnikov, A.}, \bibinfo{author}{Weissenborn, D.}, \bibinfo{author}{Zhai, X.}, \bibinfo{author}{Unterthiner, T.}, \bibinfo{author}{Dehghani, M.}, \bibinfo{author}{Minderer, M.}, \bibinfo{author}{Heigold, G.}, \bibinfo{author}{Gelly, S.}, et~al., \bibinfo{year}{2020}.
\newblock \bibinfo{title}{An image is worth 16x16 words: Transformers for image recognition at scale}.
\newblock \bibinfo{journal}{arXiv preprint arXiv:2010.11929} .
%Type = Article
\bibitem[{Eitz et~al.(2012)Eitz, Hays and Alexa}]{10.1145/2185520.2185540}
\bibinfo{author}{Eitz, M.}, \bibinfo{author}{Hays, J.}, \bibinfo{author}{Alexa, M.}, \bibinfo{year}{2012}.
\newblock \bibinfo{title}{How do humans sketch objects?}
\newblock \bibinfo{journal}{ACM Trans. Graph.} \bibinfo{volume}{31}.
\newblock \URLprefix \url{https://doi.org/10.1145/2185520.2185540}, \DOIprefix\doi{10.1145/2185520.2185540}.
%Type = Article
\bibitem[{Eitz et~al.(2010)Eitz, Hildebrand, Boubekeur and Alexa}]{hog_svm}
\bibinfo{author}{Eitz, M.}, \bibinfo{author}{Hildebrand, K.}, \bibinfo{author}{Boubekeur, T.}, \bibinfo{author}{Alexa, M.}, \bibinfo{year}{2010}.
\newblock \bibinfo{title}{Sketch-based image retrieval: Benchmark and bag-of-features descriptors}.
\newblock \bibinfo{journal}{IEEE transactions on visualization and computer graphics} \bibinfo{volume}{17}, \bibinfo{pages}{1624--1636}.
%Type = Article
\bibitem[{Ge et~al.(2020)Ge, Goswami, Zitnick and Parikh}]{ge2021creativesketchgeneration}
\bibinfo{author}{Ge, S.}, \bibinfo{author}{Goswami, V.}, \bibinfo{author}{Zitnick, C.L.}, \bibinfo{author}{Parikh, D.}, \bibinfo{year}{2020}.
\newblock \bibinfo{title}{Creative sketch generation}.
\newblock \bibinfo{journal}{arXiv preprint arXiv:2011.10039} .
%Type = Article
\bibitem[{Guo et~al.(2016)Guo, Wang, Roman-Rangel, Chao and Rui}]{7327196}
\bibinfo{author}{Guo, J.}, \bibinfo{author}{Wang, C.}, \bibinfo{author}{Roman-Rangel, E.}, \bibinfo{author}{Chao, H.}, \bibinfo{author}{Rui, Y.}, \bibinfo{year}{2016}.
\newblock \bibinfo{title}{Building hierarchical representations for oracle character and sketch recognition}.
\newblock \bibinfo{journal}{IEEE Transactions on Image Processing} \bibinfo{volume}{25}, \bibinfo{pages}{104--118}.
\newblock \DOIprefix\doi{10.1109/TIP.2015.2500019}.
%Type = Inproceedings
\bibitem[{Ha and Eck(2018)}]{ha2017neuralrepresentationsketchdrawings}
\bibinfo{author}{Ha, D.}, \bibinfo{author}{Eck, D.}, \bibinfo{year}{2018}.
\newblock \bibinfo{title}{A neural representation of sketch drawings}, in: \bibinfo{booktitle}{International Conference on Learning Representations}.
\newblock \URLprefix \url{https://openreview.net/forum?id=Hy6GHpkCW}.
%Type = Inproceedings
\bibitem[{He et~al.(2016)He, Zhang, Ren and Sun}]{resnet}
\bibinfo{author}{He, K.}, \bibinfo{author}{Zhang, X.}, \bibinfo{author}{Ren, S.}, \bibinfo{author}{Sun, J.}, \bibinfo{year}{2016}.
\newblock \bibinfo{title}{Deep residual learning for image recognition}, in: \bibinfo{booktitle}{Proceedings of the IEEE conference on computer vision and pattern recognition}, pp. \bibinfo{pages}{770--778}.
%Type = Article
\bibitem[{Ho et~al.(2020)Ho, Jain and Abbeel}]{ho2020denoising}
\bibinfo{author}{Ho, J.}, \bibinfo{author}{Jain, A.}, \bibinfo{author}{Abbeel, P.}, \bibinfo{year}{2020}.
\newblock \bibinfo{title}{Denoising diffusion probabilistic models}.
\newblock \bibinfo{journal}{Advances in neural information processing systems} \bibinfo{volume}{33}, \bibinfo{pages}{6840--6851}.
%Type = Article
\bibitem[{Hochreiter and Schmidhuber(1997)}]{bilstm}
\bibinfo{author}{Hochreiter, S.}, \bibinfo{author}{Schmidhuber, J.}, \bibinfo{year}{1997}.
\newblock \bibinfo{title}{Long short-term memory}.
\newblock \bibinfo{journal}{Neural computation} \bibinfo{volume}{9}, \bibinfo{pages}{1735--1780}.
%Type = Inproceedings
\bibitem[{Hu et~al.(2018)Hu, Li, Song, Xiang and Hospedales}]{hu2018sketchaclassifiersketchbasedphotoclassifier}
\bibinfo{author}{Hu, C.}, \bibinfo{author}{Li, D.}, \bibinfo{author}{Song, Y.Z.}, \bibinfo{author}{Xiang, T.}, \bibinfo{author}{Hospedales, T.M.}, \bibinfo{year}{2018}.
\newblock \bibinfo{title}{Sketch-a-classifier: Sketch-based photo classifier generation}, in: \bibinfo{booktitle}{Proceedings of the IEEE Conference on Computer Vision and Pattern Recognition (CVPR)}.
%Type = Inproceedings
\bibitem[{Jia et~al.(2020)Jia, Fan, Yu, Liu, Wang and Latecki}]{10.1145/3394171.3413810}
\bibinfo{author}{Jia, Q.}, \bibinfo{author}{Fan, X.}, \bibinfo{author}{Yu, M.}, \bibinfo{author}{Liu, Y.}, \bibinfo{author}{Wang, D.}, \bibinfo{author}{Latecki, L.J.}, \bibinfo{year}{2020}.
\newblock \bibinfo{title}{Coupling deep textural and shape features for sketch recognition}, in: \bibinfo{booktitle}{Proceedings of the 28th ACM International Conference on Multimedia}, \bibinfo{publisher}{Association for Computing Machinery}, \bibinfo{address}{New York, NY, USA}. p. \bibinfo{pages}{421–429}.
\newblock \URLprefix \url{https://doi.org/10.1145/3394171.3413810}, \DOIprefix\doi{10.1145/3394171.3413810}.
%Type = Article
\bibitem[{Kaiyrbekov and Sezgin(2020)}]{Kaiyrbekov_2020}
\bibinfo{author}{Kaiyrbekov, K.}, \bibinfo{author}{Sezgin, M.}, \bibinfo{year}{2020}.
\newblock \bibinfo{title}{Deep stroke-based sketched symbol reconstruction and segmentation}.
\newblock \bibinfo{journal}{IEEE Computer Graphics and Applications} \bibinfo{volume}{40}, \bibinfo{pages}{112--126}.
\newblock \DOIprefix\doi{10.1109/MCG.2019.2943333}.
%Type = Inproceedings
\bibitem[{Kim et~al.(2018)Kim, Wang, {\"O}ztireli and Gross}]{https://doi.org/10.1111/cgf.13365}
\bibinfo{author}{Kim, B.}, \bibinfo{author}{Wang, O.}, \bibinfo{author}{{\"O}ztireli, A.C.}, \bibinfo{author}{Gross, M.}, \bibinfo{year}{2018}.
\newblock \bibinfo{title}{Semantic segmentation for line drawing vectorization using neural networks}, in: \bibinfo{booktitle}{Computer Graphics Forum}, \bibinfo{organization}{Wiley Online Library}. pp. \bibinfo{pages}{329--338}.
%Type = Inproceedings
\bibitem[{Li et~al.(2020)Li, Gao, Shen, Zhang, Mei and Ren}]{li2020sketchman}
\bibinfo{author}{Li, J.}, \bibinfo{author}{Gao, N.}, \bibinfo{author}{Shen, T.}, \bibinfo{author}{Zhang, W.}, \bibinfo{author}{Mei, T.}, \bibinfo{author}{Ren, H.}, \bibinfo{year}{2020}.
\newblock \bibinfo{title}{Sketchman: Learning to create professional sketches}, in: \bibinfo{booktitle}{Proceedings of the 28th ACM international conference on multimedia}, pp. \bibinfo{pages}{3237--3245}.
%Type = Inproceedings
\bibitem[{Li et~al.(2013)Li, Song, Gong et~al.}]{ensemble}
\bibinfo{author}{Li, Y.}, \bibinfo{author}{Song, Y.Z.}, \bibinfo{author}{Gong, S.}, et~al., \bibinfo{year}{2013}.
\newblock \bibinfo{title}{Sketch recognition by ensemble matching of structured features.}, in: \bibinfo{booktitle}{BMVC}, p.~\bibinfo{pages}{2}.
%Type = Inproceedings
\bibitem[{Lin et~al.(2019)Lin, Fu, Lu, Gong, Xue and Jiang}]{tcnet}
\bibinfo{author}{Lin, H.}, \bibinfo{author}{Fu, Y.}, \bibinfo{author}{Lu, P.}, \bibinfo{author}{Gong, S.}, \bibinfo{author}{Xue, X.}, \bibinfo{author}{Jiang, Y.G.}, \bibinfo{year}{2019}.
\newblock \bibinfo{title}{Tc-net for isbir: Triplet classification network for instance-level sketch based image retrieval}, in: \bibinfo{booktitle}{Proceedings of the 27th ACM international conference on multimedia}, pp. \bibinfo{pages}{1676--1684}.
%Type = Inproceedings
\bibitem[{Lin et~al.(2020)Lin, Fu, Xue and Jiang}]{lin2020sketchbertlearningsketchbidirectional}
\bibinfo{author}{Lin, H.}, \bibinfo{author}{Fu, Y.}, \bibinfo{author}{Xue, X.}, \bibinfo{author}{Jiang, Y.G.}, \bibinfo{year}{2020}.
\newblock \bibinfo{title}{Sketch-bert: Learning sketch bidirectional encoder representation from transformers by self-supervised learning of sketch gestalt}, in: \bibinfo{booktitle}{Proceedings of the IEEE/CVF Conference on Computer Vision and Pattern Recognition (CVPR)}.
%Type = Article
\bibitem[{Lowe(2004)}]{Lowe2004DistinctiveIF}
\bibinfo{author}{Lowe, D.G.}, \bibinfo{year}{2004}.
\newblock \bibinfo{title}{Distinctive image features from scale-invariant keypoints}.
\newblock \bibinfo{journal}{International journal of computer vision} \bibinfo{volume}{60}, \bibinfo{pages}{91--110}.
%Type = Inproceedings
\bibitem[{Muhammad et~al.(2018)Muhammad, Yang, Song, Xiang and Hospedales}]{muhammad2018learningdeepsketchabstraction}
\bibinfo{author}{Muhammad, U.R.}, \bibinfo{author}{Yang, Y.}, \bibinfo{author}{Song, Y.Z.}, \bibinfo{author}{Xiang, T.}, \bibinfo{author}{Hospedales, T.M.}, \bibinfo{year}{2018}.
\newblock \bibinfo{title}{Learning deep sketch abstraction}, in: \bibinfo{booktitle}{Proceedings of the IEEE Conference on Computer Vision and Pattern Recognition (CVPR)}.
%Type = Article
\bibitem[{Qi and Tan(2019)}]{8766108}
\bibinfo{author}{Qi, Y.}, \bibinfo{author}{Tan, Z.H.}, \bibinfo{year}{2019}.
\newblock \bibinfo{title}{Sketchsegnet+: An end-to-end learning of rnn for multi-class sketch semantic segmentation}.
\newblock \bibinfo{journal}{IEEE Access} \bibinfo{volume}{7}, \bibinfo{pages}{102717--102726}.
\newblock \DOIprefix\doi{10.1109/ACCESS.2019.2929804}.
%Type = Inproceedings
\bibitem[{Ribeiro et~al.(2020)Ribeiro, Bui, Collomosse and Ponti}]{ribeiro2020sketchformertransformerbasedrepresentationsketched}
\bibinfo{author}{Ribeiro, L.S.F.}, \bibinfo{author}{Bui, T.}, \bibinfo{author}{Collomosse, J.}, \bibinfo{author}{Ponti, M.}, \bibinfo{year}{2020}.
\newblock \bibinfo{title}{Sketchformer: Transformer-based representation for sketched structure}, in: \bibinfo{booktitle}{Proceedings of the IEEE/CVF conference on computer vision and pattern recognition}, pp. \bibinfo{pages}{14153--14162}.
%Type = Article
\bibitem[{Sasaki and Ogata(2018)}]{8453841}
\bibinfo{author}{Sasaki, K.}, \bibinfo{author}{Ogata, T.}, \bibinfo{year}{2018}.
\newblock \bibinfo{title}{Adaptive drawing behavior by visuomotor learning using recurrent neural networks}.
\newblock \bibinfo{journal}{IEEE Transactions on Cognitive and Developmental Systems} \bibinfo{volume}{11}, \bibinfo{pages}{119--128}.
%Type = Inproceedings
\bibitem[{Seddati et~al.(2015)Seddati, Dupont and Mahmoudi}]{7153606}
\bibinfo{author}{Seddati, O.}, \bibinfo{author}{Dupont, S.}, \bibinfo{author}{Mahmoudi, S.}, \bibinfo{year}{2015}.
\newblock \bibinfo{title}{Deepsketch: Deep convolutional neural networks for sketch recognition and similarity search}, in: \bibinfo{booktitle}{2015 13th International Workshop on Content-Based Multimedia Indexing (CBMI)}, pp. \bibinfo{pages}{1--6}.
\newblock \DOIprefix\doi{10.1109/CBMI.2015.7153606}.
%Type = Inproceedings
\bibitem[{Seddati et~al.(2016)Seddati, Dupont and Mahmoudi}]{10.1145/2964284.2973828}
\bibinfo{author}{Seddati, O.}, \bibinfo{author}{Dupont, S.}, \bibinfo{author}{Mahmoudi, S.}, \bibinfo{year}{2016}.
\newblock \bibinfo{title}{Deepsketch2image: Deep convolutional neural networks for partial sketch recognition and image retrieval}, in: \bibinfo{booktitle}{Proceedings of the 24th ACM International Conference on Multimedia}, \bibinfo{publisher}{Association for Computing Machinery}, \bibinfo{address}{New York, NY, USA}. p. \bibinfo{pages}{739–741}.
\newblock \URLprefix \url{https://doi.org/10.1145/2964284.2973828}, \DOIprefix\doi{10.1145/2964284.2973828}.
%Type = Inproceedings
\bibitem[{Song et~al.(2017)Song, Yu, Song, Xiang and Hospedales}]{dssa}
\bibinfo{author}{Song, J.}, \bibinfo{author}{Yu, Q.}, \bibinfo{author}{Song, Y.Z.}, \bibinfo{author}{Xiang, T.}, \bibinfo{author}{Hospedales, T.M.}, \bibinfo{year}{2017}.
\newblock \bibinfo{title}{Deep spatial-semantic attention for fine-grained sketch-based image retrieval}, in: \bibinfo{booktitle}{Proceedings of the IEEE international conference on computer vision}, pp. \bibinfo{pages}{5551--5560}.
%Type = Inproceedings
\bibitem[{Tiwari et~al.(2024)Tiwari, Biswas and Llad{\'o}s}]{tiwari2024sketchgptautoregressivemodelingsketch}
\bibinfo{author}{Tiwari, A.}, \bibinfo{author}{Biswas, S.}, \bibinfo{author}{Llad{\'o}s, J.}, \bibinfo{year}{2024}.
\newblock \bibinfo{title}{Sketchgpt: Autoregressive modeling for sketch generation and recognition}, in: \bibinfo{booktitle}{International Conference on Document Analysis and Recognition}, \bibinfo{organization}{Springer}. pp. \bibinfo{pages}{421--438}.
%Type = Article
\bibitem[{Van Den~Oord et~al.(2017)Van Den~Oord, Vinyals et~al.}]{oord2018neuraldiscreterepresentationlearning}
\bibinfo{author}{Van Den~Oord, A.}, \bibinfo{author}{Vinyals, O.}, et~al., \bibinfo{year}{2017}.
\newblock \bibinfo{title}{Neural discrete representation learning}.
\newblock \bibinfo{journal}{Advances in neural information processing systems} \bibinfo{volume}{30}.
%Type = Article
\bibitem[{Vaswani et~al.(2017)Vaswani, Shazeer, Parmar, Uszkoreit, Jones, Gomez, Kaiser and Polosukhin}]{vaswani2023attentionneed}
\bibinfo{author}{Vaswani, A.}, \bibinfo{author}{Shazeer, N.}, \bibinfo{author}{Parmar, N.}, \bibinfo{author}{Uszkoreit, J.}, \bibinfo{author}{Jones, L.}, \bibinfo{author}{Gomez, A.N.}, \bibinfo{author}{Kaiser, {\L}.}, \bibinfo{author}{Polosukhin, I.}, \bibinfo{year}{2017}.
\newblock \bibinfo{title}{Attention is all you need}.
\newblock \bibinfo{journal}{Advances in neural information processing systems} \bibinfo{volume}{30}.
%Type = Inproceedings
\bibitem[{Wang and Li(2015)}]{7351724}
\bibinfo{author}{Wang, F.}, \bibinfo{author}{Li, Y.}, \bibinfo{year}{2015}.
\newblock \bibinfo{title}{Spatial matching of sketches without point correspondence}, in: \bibinfo{booktitle}{2015 IEEE International Conference on Image Processing (ICIP)}, pp. \bibinfo{pages}{4828--4832}.
\newblock \DOIprefix\doi{10.1109/ICIP.2015.7351724}.
%Type = Inproceedings
\bibitem[{Wang et~al.(2019)Wang, Lin, Wu, Li, Wang, Luo and He}]{SPFusionNet}
\bibinfo{author}{Wang, F.}, \bibinfo{author}{Lin, S.}, \bibinfo{author}{Wu, H.}, \bibinfo{author}{Li, H.}, \bibinfo{author}{Wang, R.}, \bibinfo{author}{Luo, X.}, \bibinfo{author}{He, X.}, \bibinfo{year}{2019}.
\newblock \bibinfo{title}{Spfusionnet: Sketch segmentation using multi-modal data fusion}, in: \bibinfo{booktitle}{2019 IEEE International Conference on Multimedia and Expo (ICME)}, pp. \bibinfo{pages}{1654--1659}.
\newblock \DOIprefix\doi{10.1109/ICME.2019.00285}.
%Type = Inproceedings
\bibitem[{Wang and Li(2024)}]{Wang_2024_CVPR}
\bibinfo{author}{Wang, J.}, \bibinfo{author}{Li, C.}, \bibinfo{year}{2024}.
\newblock \bibinfo{title}{Contextseg: Sketch semantic segmentation by querying the context with attention}, in: \bibinfo{booktitle}{Proceedings of the IEEE/CVF Conference on Computer Vision and Pattern Recognition (CVPR)}, pp. \bibinfo{pages}{3679--3688}.
%Type = Inproceedings
\bibitem[{Wu et~al.(2018)Wu, Qi, Liu and Yang}]{Sketchsegnet}
\bibinfo{author}{Wu, X.}, \bibinfo{author}{Qi, Y.}, \bibinfo{author}{Liu, J.}, \bibinfo{author}{Yang, J.}, \bibinfo{year}{2018}.
\newblock \bibinfo{title}{Sketchsegnet: A rnn model for labeling sketch strokes}, in: \bibinfo{booktitle}{2018 IEEE 28th International Workshop on Machine Learning for Signal Processing (MLSP)}, \bibinfo{organization}{IEEE}. pp. \bibinfo{pages}{1--6}.
%Type = Inproceedings
\bibitem[{Xu et~al.(2018)Xu, Huang, Yuan, Pang, Song, Xiang, Hospedales, Ma and Guo}]{xu2018sketchmatedeephashingmillionscale}
\bibinfo{author}{Xu, P.}, \bibinfo{author}{Huang, Y.}, \bibinfo{author}{Yuan, T.}, \bibinfo{author}{Pang, K.}, \bibinfo{author}{Song, Y.Z.}, \bibinfo{author}{Xiang, T.}, \bibinfo{author}{Hospedales, T.M.}, \bibinfo{author}{Ma, Z.}, \bibinfo{author}{Guo, J.}, \bibinfo{year}{2018}.
\newblock \bibinfo{title}{Sketchmate: Deep hashing for million-scale human sketch retrieval}, in: \bibinfo{booktitle}{Proceedings of the IEEE Conference on Computer Vision and Pattern Recognition (CVPR)}.
%Type = Article
\bibitem[{Xu et~al.(2022)Xu, Joshi and Bresson}]{xu2021multigraphtransformerfreehandsketch}
\bibinfo{author}{Xu, P.}, \bibinfo{author}{Joshi, C.K.}, \bibinfo{author}{Bresson, X.}, \bibinfo{year}{2022}.
\newblock \bibinfo{title}{Multigraph transformer for free-hand sketch recognition}.
\newblock \bibinfo{journal}{IEEE Transactions on Neural Networks and Learning Systems} \bibinfo{volume}{33}, \bibinfo{pages}{5150--5161}.
\newblock \DOIprefix\doi{10.1109/TNNLS.2021.3069230}.
%Type = Article
\bibitem[{Yang et~al.(2021)Yang, Zhuang, Fu, Wei, Zhou and Zheng}]{10.1145/3450284}
\bibinfo{author}{Yang, L.}, \bibinfo{author}{Zhuang, J.}, \bibinfo{author}{Fu, H.}, \bibinfo{author}{Wei, X.}, \bibinfo{author}{Zhou, K.}, \bibinfo{author}{Zheng, Y.}, \bibinfo{year}{2021}.
\newblock \bibinfo{title}{Sketchgnn: Semantic sketch segmentation with graph neural networks}.
\newblock \bibinfo{journal}{ACM Trans. Graph.} \bibinfo{volume}{40}.
\newblock \URLprefix \url{https://doi.org/10.1145/3450284}, \DOIprefix\doi{10.1145/3450284}.
%Type = Inproceedings
\bibitem[{Yu et~al.(2016)Yu, Liu, Song, Xiang, Hospedales and Loy}]{7780462}
\bibinfo{author}{Yu, Q.}, \bibinfo{author}{Liu, F.}, \bibinfo{author}{Song, Y.Z.}, \bibinfo{author}{Xiang, T.}, \bibinfo{author}{Hospedales, T.M.}, \bibinfo{author}{Loy, C.C.}, \bibinfo{year}{2016}.
\newblock \bibinfo{title}{Sketch me that shoe}, in: \bibinfo{booktitle}{Proceedings of the IEEE Conference on Computer Vision and Pattern Recognition (CVPR)}.
%Type = Inproceedings
\bibitem[{Yu et~al.(2015)Yu, Yang, Song, Xiang and Hospedales}]{yu2015sketchanetbeatshumans}
\bibinfo{author}{Yu, Q.}, \bibinfo{author}{Yang, Y.}, \bibinfo{author}{Song, Y.Z.}, \bibinfo{author}{Xiang, T.}, \bibinfo{author}{Hospedales, T.M.}, \bibinfo{year}{2015}.
\newblock \bibinfo{title}{Sketch-a-net that beats humans}, in: \bibinfo{booktitle}{British Machine Vision Conference}.
\newblock \URLprefix \url{https://api.semanticscholar.org/CorpusID:15004083}.
%Type = Inproceedings
\bibitem[{Zhang et~al.(2016a)Zhang, Liu, Zhang, Ren, Wang and Cao}]{7780494}
\bibinfo{author}{Zhang, H.}, \bibinfo{author}{Liu, S.}, \bibinfo{author}{Zhang, C.}, \bibinfo{author}{Ren, W.}, \bibinfo{author}{Wang, R.}, \bibinfo{author}{Cao, X.}, \bibinfo{year}{2016}a.
\newblock \bibinfo{title}{Sketchnet: Sketch classification with web images}, in: \bibinfo{booktitle}{Proceedings of the IEEE Conference on Computer Vision and Pattern Recognition (CVPR)}.
%Type = Article
\bibitem[{Zhang et~al.(2019)Zhang, She, Liu, Gan, Cao and Foroosh}]{8694004}
\bibinfo{author}{Zhang, H.}, \bibinfo{author}{She, P.}, \bibinfo{author}{Liu, Y.}, \bibinfo{author}{Gan, J.}, \bibinfo{author}{Cao, X.}, \bibinfo{author}{Foroosh, H.}, \bibinfo{year}{2019}.
\newblock \bibinfo{title}{Learning structural representations via dynamic object landmarks discovery for sketch recognition and retrieval}.
\newblock \bibinfo{journal}{IEEE Transactions on Image Processing} \bibinfo{volume}{28}, \bibinfo{pages}{4486--4499}.
\newblock \DOIprefix\doi{10.1109/TIP.2019.2910398}.
%Type = Inproceedings
\bibitem[{Zhang et~al.(2016b)Zhang, Zhang and Qian}]{Zhang2016DeepNN}
\bibinfo{author}{Zhang, Y.}, \bibinfo{author}{Zhang, Y.}, \bibinfo{author}{Qian, X.}, \bibinfo{year}{2016}b.
\newblock \bibinfo{title}{Deep neural networks for free-hand sketch recognition}, in: \bibinfo{booktitle}{Advances in Multimedia Information Processing-PCM 2016: 17th Pacific-Rim Conference on Multimedia, Xi{\'{}} an, China, September 15-16, 2016, Proceedings, Part II}, \bibinfo{organization}{Springer}. pp. \bibinfo{pages}{689--696}.

\end{thebibliography}
